\documentclass[letterpaper, 10 pt, conference]{ieeeconf}  

\IEEEoverridecommandlockouts                              
\usepackage{amsmath}
\usepackage{graphicx}
\usepackage{caption}
\usepackage{xcolor}
\usepackage{cuted}
\usepackage{svg}
\usepackage{multirow}
\usepackage{hyperref}
\usepackage{balance}
\title{\LARGE \bf
Embodiment-Aware Planning for Heterogeneous Bimanual Manipulation
}

\author{Sukhvansh Jain$^{1}$, Anurag Maurya$^{2}$, Shivam Vats$^{3}$, Narendra Kumar Dhar$^{1}$, and Ravi Prakash$^{2}$%
\thanks{$^{1}$Sukhvansh Jain and Narendra Kumar Dhar are with the Indian Institute of Technology Mandi, Mandi, HP, India. (Corresponding author: {\tt\small sukhvanshjain@gmail.com})}%
\thanks{$^{2}$Anurag Maurya and Ravi Prakash are with the Indian Institute of Science, Bengaluru, Karnataka, India.}%
\thanks{$^{3}$Shivam Vats is with Brown University, Providence, RI, USA.}%
}

\begin{document}

\maketitle

\thispagestyle{empty}
\pagestyle{empty}

\begin{strip}
\vspace{-15ex}
\centering
\includegraphics[width=\textwidth]{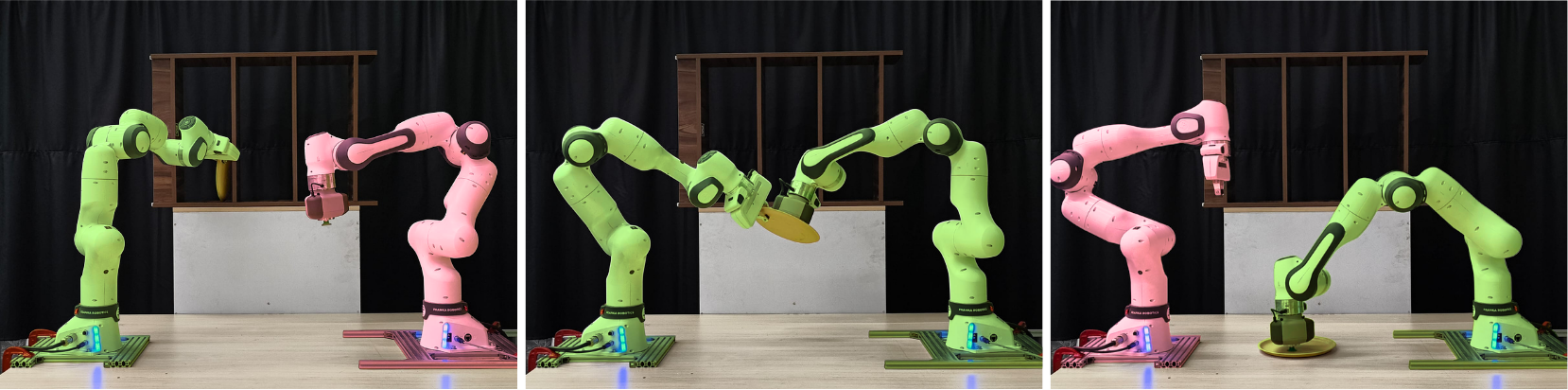}
\captionof{figure}{\textbf{Heterogeneous Gripper coordination for Bimanual Manipulation} A plate transfer task that is infeasible for either gripper alone. The parallel-jaw gripper cannot safely perform the placement due to collisions with table, while the suction gripper cannot complete the constrained shelf pickup. By coordinating complementary simple grippers, the system extends its manipulation capability: parallel jaw performs the initial pick and transport (left), a dual-arm handover enables modality transition (center), and the suction gripper completes precise placement (right). Active arm(s) are shown in green, inactive in red. This example illustrates how heterogeneous coordination can approximate the versatility of human dexterity using simple industrial hardware.}
\label{fig:teaser}
\end{strip}

\begin{abstract}

Many manipulation tasks that are effortless for dexterous human hands remain challenging for robotic systems equipped with simple, task-specialized grippers. Industrial platforms commonly rely on end-effectors such as suction cups or parallel-jaw grippers, yet no single mode can reliably satisfy the diverse constraints encountered across different phases of complex manipulation tasks, including pickup, reorientation, insertion, and constrained placement.
In this work, we investigate whether complementary simple grippers within a heterogeneous bimanual setup can collectively realize tasks that exceed the capability of any individual gripper alone. Given a single human demonstration specifying the desired object motion, we formulate embodiment-aware execution as a finite-horizon optimization over arm–trajectory assignments. At each segment, the planner determines whether single-arm manipulation or a coordinated dual-arm handover is required, jointly reasoning over heterogeneous grasp feasibility, motion constraints, manipulability, and cross-gripper compatibility. Exploiting this formulation, we compute globally consistent assignments with explicit switching regularization.
Experiments on a dual Franka FR3 platform equipped with suction and parallel-jaw grippers demonstrate improved feasibility and reliability in tasks that are infeasible under single-gripper execution. \href{https://sukhvansh2004.github.io/Bimanual-Manipulation/}{\textbf{WEBSITE}}

\end{abstract}


\section{Introduction}
Humans perform complex manipulation tasks using highly dexterous hands capable of continuously varying contact modes, regrasping strategies, and in-hand adjustments. In contrast, most robotic systems rely on simple, task-specialized end-effectors such as suction cups or parallel-jaw grippers. While these grippers are robust, affordable, and widely deployed in industrial environments, each offers limited affordances. A suction gripper excels at lifting flat surfaces but struggles in confined insertions; a parallel-jaw gripper provides stable force-closure grasps yet often fails on thin or flush objects. Consequently, many tasks that are effortless for human hands cannot be reliably executed using a single simple robotic gripper.
\\

In modern industrial settings such as warehouse automation, packaging lines, and retail fulfillment, objects vary widely in geometry, thickness, surface properties, and accessibility. During a single transport task, different phases may demand different grasp modalities. Rather than relying on complex anthropomorphic hands, a practical alternative is to combine complementary simple grippers within a heterogeneous bimanual system. Such a configuration has the potential to approximate the functional diversity of human manipulation while retaining the robustness and simplicity of industrial hardware. However, enabling two specialized grippers to collectively realize tasks that exceed the capability of either alone requires systematic coordination and planning.
\\

Figure~\ref{fig:teaser} illustrates this principle. A thin plate must be transferred from a narrow shelf to a table. The suction gripper cannot safely execute the pickup due to geometric interference during approach, whereas the parallel jaw gripper successfully performs the initial pick and transport. However, it cannot complete the placement due to collision constraints with table. The task becomes feasible only when the system performs a coordinated dual-arm handover, allowing the suction gripper to execute the final placement. Importantly, neither gripper alone can solve the task, yet their coordinated use enables seamless execution. This example captures the central premise of our work: heterogeneous coordination can elevate simple industrial grippers' performance, provided that switching and compatibility are planned systematically.
\\

In this work, we study embodiment-aware realization of demonstrated manipulation tasks using a heterogeneous dual-arm robotic system. A single human demonstration provides the desired object motion as a task-level specification. Rather than learning a policy, our objective is to determine how the demonstrated task can be executed under the constraints of heterogeneous robotic hardware. Given a discretized object trajectory, we determine at each segment which arm should manipulate the object and when coordinated handovers are necessary to maintain feasibility under grasp, kinematic, and collision constraints.
\\

Heterogeneous execution introduces unique planning challenges. At different phases of a task, only specific gripper modalities may satisfy feasibility requirements. Coordinated handovers must respect cross-gripper grasp compatibility, and excessive switching increases execution time and failure risk creating the need for a global reasoning. We formulate this problem as a finite-horizon optimization over arm–trajectory assignments. Each state encodes both the active robot and its associated grasp-induced motion. Segment costs capture feasibility and manipulability quality, while transition costs encode heterogeneous grasp-pair compatibility and switching penalties. Exploiting the optimal substructure of this formulation, we compute a globally consistent execution plan via dynamic programming, equivalent to a structured shortest-path computation over a time-expanded graph.
\\

Our contributions are threefold:
\begin{enumerate}
    \item \textbf{Heterogeneous gripper-level coordination:} We formalize how a complementary simple gripper system can collectively realize manipulation tasks that are infeasible for any single modality alone, extending beyond simple pick-only tasks.
    \item  \textbf{Globally consistent embodiment-aware assignment:} We introduce a structured planning formulation that integrates heterogeneous grasp affordances and handover constraints into a globally consistent assignment.
    \item  \textbf{Sim-to-real transfer:} We demonstrate the framework on a dual Franka FR3 platform equipped with suction and parallel-jaw grippers on a task requiring modality transitions and coordinated handovers.
\end{enumerate}

\section{Related Work}
\subsection{Dexterous Manipulation vs. Simple Grippers}
Robotic dexterous hands aim to replicate the versatility of human manipulation through multi-fingered contact, in-hand regrasping, and rich contact reasoning \cite{lin2025dexterousreview,bicchi2000graspingreview}. While highly expressive, they introduce substantial mechanical and control complexity. Classical multi-fingered manipulation established many of the analytical foundations for contact kinematics and grasp planning \cite{kerr1986multifinger,montana1988contact,cutkosky1989grasp}. Industrial systems instead rely on simple, task-specific end-effectors such as suction cups or parallel-jaw grippers for their robustness and cost efficiency. However, each supports only limited contact modalities. This leaves the challenge of expanding manipulation capability without complex anthropomorphic hardware largely open.
\\
Recent reinforcement learning has enabled impressive dexterous manipulation on anthropomorphic hands, including large-scale sim-to-real policies and reorientation tasks \cite{openai2018dactyl,rajeswaran2018dexterous,akkaya2019rubik}. We instead explore extending capability by coordinating complementary simple grippers in a heterogeneous bimanual system. Rather than increasing single-hand dexterity, we study whether structured coordination between specialized modalities can approximate the functional diversity.

\subsection{Bimanual Manipulation and Grasp Planning}

Bimanual manipulation has been widely studied for coordinated transport, reorientation, and in-hand manipulation \cite{ISLAM2026101050}. Early work largely focused on symmetric dual-arm systems with identical end-effectors, emphasizing synchronized motion planning and closed-chain control \cite{shome2018fast,MARTINEZPERAL2025105118}. Several works also address grasp generation and object handover between two arms. GraspGen \cite{murali2025graspgen} generates grasps for parallel-jaw grippers, while sampling-based and graph-based methods search over candidate grasp pairs and intermediate transfer poses \cite{handoverTSQP} \cite{Cohen}.

More recently, learning-based approaches have demonstrated coordinated dual-arm behaviors, including end-to-end Vision–Language–Action models \cite{black2024pi_0}, learned bimanual coordination \cite{chen2025bimanual}, and reactive self-handover in confined environments \cite{geng2024selfhandover}. Manipulability-aware dual-arm grasp planning has also been explored in redundant systems \cite{wong2024manipulability}, along with physics-aware saliency learning for bimanual grasp point selection \cite{wang2024bimanualsaliency}. However, most existing methods assume identical end-effectors or treat grasp feasibility separately for each arm. Coordination with heterogeneous grippers and explicit reasoning over grasp compatibility and switching across task phases still remains a challenge. Our approach instead performs global planning that jointly models grasp compatibility, motion feasibility, and modality transitions.

\subsection{Heterogeneous Gripper Systems}
There is limited prior work on planning with heterogeneous grippers to extend the manipulation capability of bimanual robots. HeterBot \cite{heterbot} introduces a mobile platform with two different grippers but focuses mainly on teleoperation rather than autonomous coordination. VIKI-R \cite{kangviki} assigns task primitives across heterogeneous platforms on high-level, but does not explicitly reason about cross-gripper low-level manipulation. \cite{app11135805},\cite{wu2025wild} and \cite{sciroboticsambidexterous} employs heterogeneous grippers; however, its applicability is restricted to object picking tasks. Overall, existing work focuses on gripper selection for individual pick actions, while holistic task planning and feasibility across heterogeneous embodiments remain understudied.
\\
\textbf{Positioning:} Most existing methods assume identical end-effectors, treat grasp selection separately from task planning, or rely on heuristic or manual switching between arms. In contrast, our method jointly reasons about grasp feasibility, grasp-pair compatibility, and arm switching in a single global planning framework.

\section{Problem Formulation}

We consider a heterogeneous bimanual manipulation system with two robot arms $\mathcal{R} := \{R_L, R_R\}$ operating in a shared workspace. Each arm $R_j$ is equipped with a possibly different end-effector $G_j \in \{\text{parallel-jaw}, \text{suction}, \dots\}$, leading to asymmetric grasp affordances.

\textbf{Task Setup:} An object $o$ must be moved from a start pose $X_{start} \in SE(3)$ to a goal pose $X_{goal} \in SE(3)$. A single human demonstration provides an object-centric trajectory $\xi : [0,T] \rightarrow SE(3)$ describing the desired object motion. The trajectory is discretized into $N$ ordered segments $\{\xi_1, \dots, \xi_N\}$. Due to heterogeneous gripper affordances and kinematic limits, different segments may only be executable by specific arms.

For each segment $\xi_i$, we construct a finite set of feasible arm–trajectory pairs $s_i = (r_i, \tau_i)$, where $r_i \in \mathcal{R}$ is the active robot and $\tau_i$ is a grasp-conditioned, collision-free trajectory that executes $\xi_i$. Let $\mathcal{S}$ denote the set of all feasible states. A plan is a sequence $\mathbf{s}_{1:N} = (s_1, \dots, s_N)$ such that exactly one arm manipulates the object at each segment. The objective is to select a sequence that remains feasible while minimizing motion and switching effort over the full task.

\section{Embodiment-Aware Planning Framework}

\begin{figure*}[t]
    \centering
    \includegraphics[width=\linewidth]{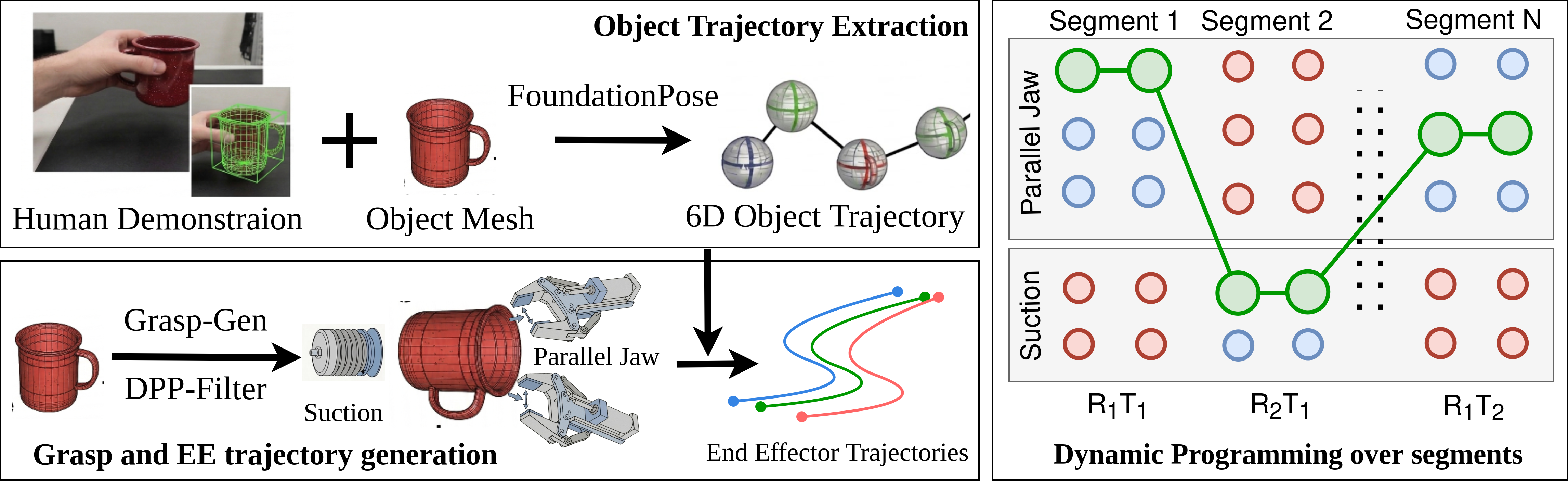}
    \caption{Overview of the planning framework. For each trajectory segment, feasible robot–trajectory pairs are generated from grasp and motion feasibility checks. Segment costs capture execution quality, while transition costs encode switching penalties and grasp compatibility. A dynamic program then computes the globally optimal sequence of arm assignments over the full task horizon. Red denotes infeasible regions, blue denotes high-cost regions, and green indicates the optimal path for the task.}
    \label{fig:methodology}
\end{figure*}

Given the feasible state set $\mathcal{S}$ for each segment, task execution is formulated as a finite-horizon discrete optimization problem. Candidate grasps on the object generate multiple trajectory instances for each arm. These candidates are evaluated for grasp validity, kinematic reachability, and collision-free motion. Each feasible state is assigned a segment cost, while transitions between states encode switching penalties and grasp compatibility. A dynamic programming procedure then computes the minimum-cost sequence of states over the task horizon, producing a feasible and smooth bimanual execution plan such as $\{(r_1, \tau_1), (r_1, \tau_1), (r_2, \tau_8), (r_2, \tau_8), \dots \}$.

The individual components of the framework are described in the following subsections.

\subsection{Trajectory Extraction from Human Demonstration}
We extract the object trajectory via model-based 6D pose tracking using \textit{FoundationPose} \cite{wen2024foundationpose}, leveraging the known CAD mesh to enforce geometric consistency during manipulation (See Figure ~\ref{fig:methodology}). While the mesh is required for planning, recent work enables reliable mesh reconstruction of novel objects from multiple RGB-D images \cite{wen2023bundlesdf}. The pipeline initializes an $SE(3)$ pose from RGB-D observations and then tracks the object by conditioning each estimate on the previous pose to maintain temporal coherence. At each timestep, pose alignment is optimized between observed depth and the rendered mesh. The resulting poses are temporally filtered to suppress high-frequency jitter while preserving rigid-body consistency, yielding smooth trajectories robust to partial occlusion. From these trajectory points, we define motion segments $\xi_i = (p_i, p_{i+1})$ between consecutive poses.

While Foundation Pose and trajectory extraction from Human Demonstration serve as a general approach, our method works with any method that can provide an object trajectory; thus, we also tested our approach with the RRT* algorithm for object trajectory generation.

\subsection{Grasp Generation, Ranking and End-Effector Trajectory Formation}
\label{Grasp-Generation}

A feasible plan requires at least one robot to have a valid grasp pose $g$ for gripper $G_j$ at object pose $p_i$. Let $\mathcal{G}_j=\texttt{SampleGrasps}(o,G_j)$ denote the candidate grasp set; feasibility requires $\mathcal{G}_j\neq\emptyset$. This captures gripper-dependent affordances (e.g., suction on flat surfaces and parallel-jaw grasping on accessible edges), modeled over the coupled state of both grippers and the object rather than any single gripper in isolation. We use a geometry-based grasp generator \cite{murali2025graspgen}. Scene-conditioned generation is avoided because a grasp infeasible in one segment may become feasible later due to changes in arm assignment or handover configuration, complicating consistent reasoning across segments.

To encourage grasp diversity, we bypass the generator’s internal ranking and instead perform entropy-based subset selection. Given candidate grasps $\{g_1,\dots,g_M\}$ with positive quality scores $w_i$, we construct a Determinantal Point Process (DPP) using an $L$-ensemble. The $L$-ensemble matrix is defined as
\begin{equation}
L_{ij} = w_i w_j K_{ij}, 
\qquad 
K_{ij} = \exp\!\left(-\tfrac{1}{2} d(g_i,g_j)^2\right).
\end{equation}
The grasp distance metric is given by
\begin{equation}
d(g_i,g_j)^2
=
\frac{\|p_{g,i} - p_{g,j}\|^2}{\sigma_p^2}
+
\frac{\bigl(1 - |\langle o_{g,i}, o_{g,j}\rangle|\bigr)^2}{\sigma_r^2}.
\end{equation}
where $p_{g,i}$ and $o_{g,i}$ denote the position and orientation (quaternion) of grasp $g_i$, respectively.
We select a subset $S$ of size $k$ by greedily maximizing 
$\log \det(L_S)$, which jointly promotes grasp quality (via $w_i$) 
and spatial--rotational diversity (via $K_{ij}$). Each selected grasp is used to generate the corresponding end-effector trajectory. These trajectories share the same time parameterization as the original object trajectory. 

\subsection{Segment-wise Feasibility and Cost Formulation}
\label{segment-feasibility}

For each transport segment $\xi_i$, we determine which arm $R_j \in \mathcal{R} = \{R_L, R_R\}$ can execute the motion. An action is defined as $A_{i,j} = (o, p_i, p_{i+1}, R_j)$, representing transport of object $o$ from $p_i$ to $p_{i+1}$ using arm $R_j$. Its feasibility $F(A_{i,j}) \in \{\text{True}, \text{False}\}$ depends on grasp and motion feasibility.

\subsubsection{Grasp Feasibility}
A valid grasp must exist for gripper $G_j$ at pose $p_i$. Let $\mathcal{G}_{i,j} = \texttt{SampleGrasps}(o, p_i, G_j)$ denote the candidate grasp set; feasibility requires $\mathcal{G}_{i,j} \neq \emptyset$ (already enforced by the filtering step in Section~\ref{Grasp-Generation}). We additionally compute mutually feasible grasp pairs across robots, i.e., collision-free grasp pairs that can simultaneously hold the object, which are required for handovers.

\subsubsection{Motion Planning Feasibility}
A collision-free trajectory must exist for arm $R_j$ to transport the object from $p_i$ to $p_{i+1}$. Feasible motions must also exist for pick at $p_0$, release at $p_T$, and handover execution. If any motion primitive is infeasible, the action is rejected.

An analytical Inverse Kinematics (EAIK) \cite{ostermeier2025automatic} solver evaluates exact endpoint reachability in microseconds, acting as a rapid feasibility filter. If analytical reachability is confirmed, we then evaluate the required motion primitives.

\subsubsection{Segment Cost}

In order to avoid awkward or near-singular configurations, we define a task cost based on the Yoshikawa manipulability measure and smoothness of the transitions between points. For a configuration $q_{i,j}$ corresponding to action $A_{i,j}$, 
we define two cost components.

\paragraph{Smoothness Cost.}
The smoothness term penalizes large joint variations between 
consecutive configurations:
\begin{equation}
C_1(A_{i,j}) =
\ \left\lVert q_{i+1,j} - q_{i,j} \right\rVert^2.
\end{equation}

\paragraph{Manipulability Cost.}
Given the Jacobian $J(q)$, the Yoshikawa manipulability index \cite{yoshikawa1985manipulability} for a joint configuration $q$ is
\begin{equation}
w(q) =
\sqrt{\det\!\left(J(q) J(q)^\top\right)}.
\end{equation}
To penalize near-singular configurations along the transport segment, we evaluate the index at both the start ($q_{i,j}$) and end ($q_{i+1,j}$) configurations of the action. We define the cost as:
\begin{equation}
C_2(A_{i,j}) =
\frac{1}{w(q_{i,j}) + \epsilon} + \frac{1}{w(q_{i+1,j}) + \epsilon},
\end{equation}
where $\epsilon > 0$ is a small constant for numerical stability.

\paragraph{Combined Task Cost.}
Let $F(A_{i,j}) \in \{\mathrm{True}, \mathrm{False}\}$ 
denote the feasibility of the action. The overall task cost is
\begin{equation}
C(A_{i,j}) =
\begin{cases}
\alpha\, C_1(A_{i,j}) + \beta\, C_2(A_{i,j}),
& \text{if } F(A_{i,j}) = \mathrm{True}, \\[8pt]
\infty,
& \text{otherwise},
\end{cases}
\end{equation}

where $\alpha, \beta > 0$ balance smoothness and manipulability. Through a grid search for $\alpha \in \{2.5, \allowbreak 5.0, \allowbreak 10.0, \allowbreak 20.0, \allowbreak 40.0, \allowbreak 50.0\}$ and $\beta \in \{0.0, \allowbreak 0.25, \allowbreak 0.5, \allowbreak 1.0, \allowbreak 2.0, \allowbreak 4.0\}$, these weights were set to $\alpha = 40$ and $\beta = 0.5$ to yield the best performance. For kinematically redundant manipulators, we include the manipulability term directly in the inverse kinematics (IK) optimization too. By doing so, the solver naturally prefers well-conditioned configurations, resulting in smooth and highly manipulable robot motions.

\subsection{Transition Design for Heterogeneous Handover}
We define the handover primitive at a point \(p_i\) as a dual-arm exchange between a transfer and a receiver arm using predefined grasp transforms. Each end-effector pose is computed by combining the object pose with the inverse of its grasp frame. The primitive consists of three stages: \textbf{Pre-alignment}, where the object is slightly offset while the receiver aligns to this intermediate pose; \textbf{Receiver approach}, where the receiver interpolates smoothly to the final grasp along the grasp normal; and \textbf{Transfer retreat}, where the original arm retreats along a fixed axis after handoff. This sequence ensures smooth, coordinated, and geometrically consistent dual-arm handovers.

Transitions between identical robot states are disallowed by assigning infinite cost to such edges. Although counterintuitive, this is necessary for feasibility: switching planners mid-trajectory requires an instantaneous release and regrasp, which may cause the object to fall. If a grasp change is required, it can be achieved through two handovers, but handover costs discourage unnecessary frequent switching unless required for feasibility.

Following Section~\ref{segment-feasibility}, we also enforce paired heterogeneous grasp feasibility by assigning infinite cost to infeasible simultaneous grasp configurations, thereby eliminating transitions corresponding to infeasible dual-arm trajectories. For feasible handover primitives, the cost is defined analogously to the segment cost in Section~\ref{segment-feasibility}. Specifically, the total cost of a handover primitive $\mathcal{H}$ is given by
\begin{equation}
C_{\text{switch}}(\mathcal{H})
=
\sum_{(i,j) \in \mathcal{H}} C(A_{i,j}),
\end{equation}
where $C(A_{i,j})$ is the combined smoothness–manipulability cost
defined previously.

The transition cost between states $s$ and $s'$ is defined as
\begin{equation}
\Phi(s', s) =
\begin{cases}
C_{\text{switch}}, 
& r(s') \neq r(s);\ \text{grasp pair feasible}, \\[6pt]
\infty, 
& r(s') \neq r(s);\ \text{grasp pair infeasible}, \\[6pt]
\infty, 
& r(s') = r(s);\ \tau(s') \neq \tau(s).
\end{cases}
\end{equation}
Here, $r(s)$ denotes the active robot, $\tau(s)$ the associated
trajectory index, and $C_{\text{switch}} > 0$ is a switching
penalty that regularizes inter-robot handovers.

\subsection{Dynamic Programming and Optimal Sequence Recovery}

The problem admits optimal substructure and is solved via dynamic programming. The value function is defined as,
\begin{equation}
V_i(s)
=
\min_{s_{1:i-1}}
\left(
\sum_{k=1}^{i} C_k(s_k)
+
\sum_{k=2}^{i} \Phi(s_{k-1}, s_k)
\right),
\end{equation}

which represents the minimum cumulative cost of reaching segment \( i \) and terminating in state \( s \).

\noindent\paragraph{Initialization.}
\begin{equation}
V_1(s) =
\begin{cases}
C_1(s), & s \in \mathcal{S}_0, \\
\infty, & \text{otherwise}.
\end{cases}
\end{equation}
\noindent\paragraph{Bellman Recursion.}
For \( i = 2,\dots,N \), recursion is
\begin{equation}
V_i(s)
=
C_i(s)
+
\min_{s' \in \mathcal{S}}
\left(
V_{i-1}(s') + \Phi(s', s)
\right).
\end{equation}
Infeasible transitions are excluded by setting \( \Phi(s', s) = \infty \). The recursion is equivalent to a shortest-path computation on a layered directed acyclic graph.
For policy recovery, we store the predecessor mapping
\begin{equation}
\pi_i(s)
=
\arg\min_{s' \in \mathcal{S}}
\left(
V_{i-1}(s') + \Phi(s', s)
\right).
\end{equation}
Each optimal state encodes both the selected robot and its associated trajectory, i.e., 
$s_i^\star = (r_i^\star, \tau_i^\star)$. The resulting assignment sequence is 
$\mathbf{s}_{1:N}^\star = ((r_1^\star,\tau_1^\star), \dots, (r_N^\star,\tau_N^\star))$, 
from which the robot and trajectory sequences can be extracted as 
$\mathbf{r}_{1:N}^\star = (r_1^\star,\dots,r_N^\star)$ and 
$\boldsymbol{\tau}_{1:N}^\star = (\tau_1^\star,\dots,\tau_N^\star)$. 
Switching events are defined as 
$\mathcal{H} = \{\, i \in \{2,\dots,N\} \mid r(s_i^\star) \neq r(s_{i-1}^\star) \,\}$.

\subsection{Complexity Reduction via Beam Search and Stride}

To mitigate the $\mathcal{O}(N |\mathcal{S}|^2)$ worst-case time complexity of the dynamic programming pass and ensure scalability to longer horizons, we introduce a beam search with a stride parameter. We evaluate potential dual-arm handovers only every $k$ steps (stride) rather than at every discretized trajectory waypoint. Furthermore, during the forward pass, we retain only the top $B$ lowest-cost states (beam width) for evaluation at the next step. This significantly prunes the state space and branching factor, preserving performance while drastically reducing planning times.

\section{Experiments and Results}
\label{sec:experiments}

\begin{figure*}[t]
    \centering
    \includegraphics[width=0.95\linewidth]{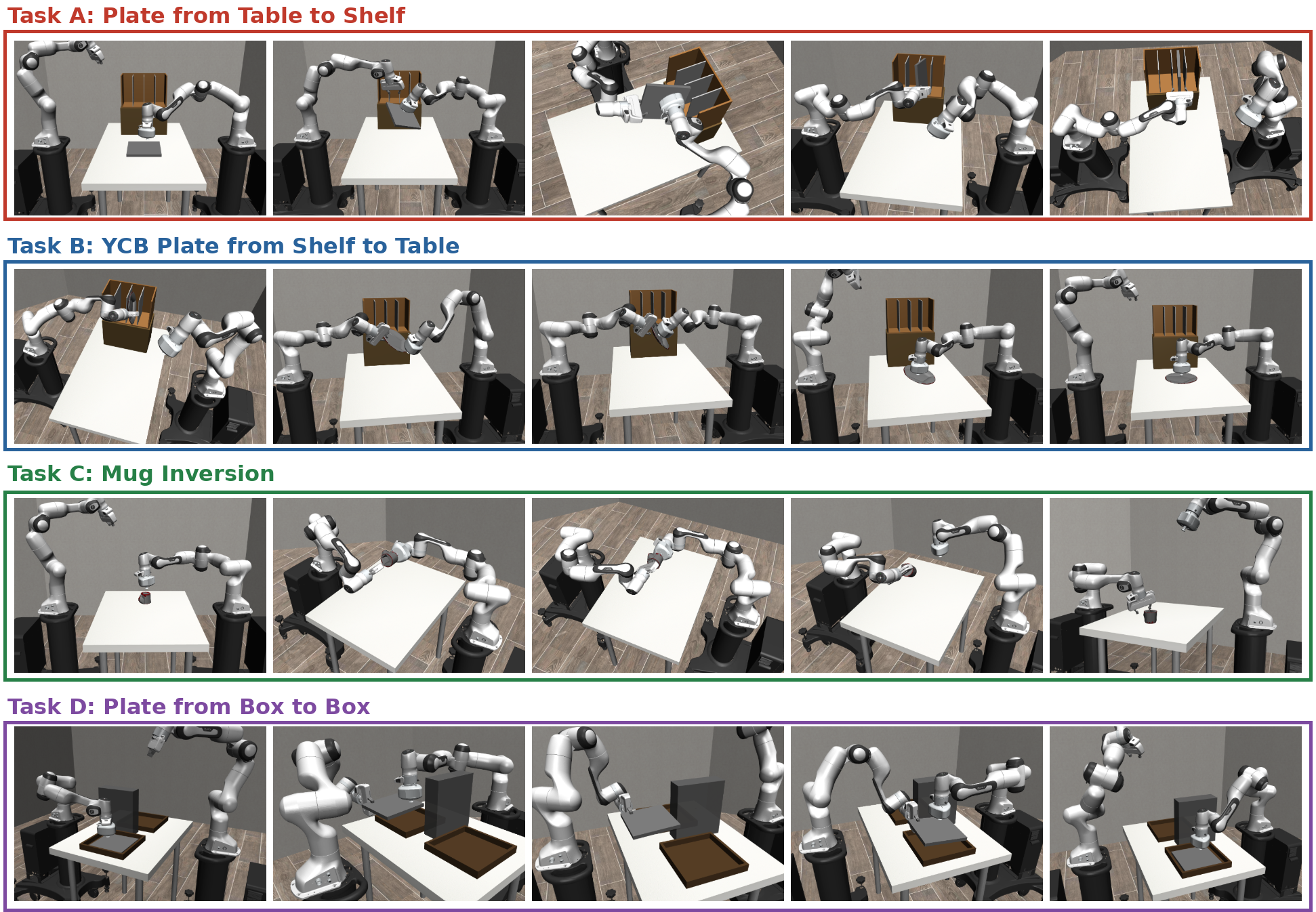}
    \caption{The figure illustrates the four tasks evaluated in our framework. The tasks are specifically designed to highlight the advantages and coordination challenges of the proposed heterogeneous bimanual setup during real-world experiments. The execution sequence of the tasks is from left to right.}
    \label{fig:tasks}
\end{figure*}

Our experiments are designed to systematically evaluate the proposed system framework. Specifically, we aim to answer the following questions:
\begin{itemize}
    \item[\textbf{Q1:}] To what extent does the proposed dynamic programming formulation improve task completion rates in scenarios involving heterogeneous grippers?
    \item[\textbf{Q2:}] What is the impact of incorporating diversity-aware grasp selection and manipulability-based costs on overall system performance?
    \item[\textbf{Q3:}] Which components are most crucial to the method’s performance, and what are the dominant failure modes observed in practice?
\end{itemize}

\subsection{Experimental Setup}
\label{sec:exp_setup}
All experiments are conducted on a heterogeneous dual-arm platform with two 7-DoF Franka FR3 manipulators: one equipped with a parallel-jaw gripper and the other with a suction gripper. The arms are arranged with overlapping workspaces to enable coordinated manipulation and object handovers, and their relative configuration is varied to induce different reachability and feasibility constraints. This setup allows evaluation of grasp-dependent affordances and asymmetric reachability. All experiments are conducted on a workstation equipped with an AMD Ryzen Threadripper PRO 7975WX processor and an NVIDIA RTX 6000 Ada Generation GPU.

\textbf{Grasp Generation, Motion Planning and Execution:}
\begin{itemize}
    \item \textbf{Inverse Kinematics:} An \texttt{EAIK} \cite{ostermeier2025automatic} based IK solver with multi-joint enumeration (using the redundancy of 7 DOF) to check for the most continuous and manipulable joint pose.
    \item \textbf{Grasp Generation:} 6-DoF grasp candidates are sampled via \textbf{GraspGen}\cite{murali2025graspgen} for both parallel-jaw and suction grippers, followed by collision-based feasibility filtering. 40 grasps are sampled per gripper type, except in Task A: Plate from Table to Shelf (as described below) due to its restrictive environment, thus requiring more diverse sampling (60 grasps).
    \item \textbf{Collision modeling:} Full robot self-collision, object and environment meshes, and dual-arm interactions checked using MuJoCo’s default collision engine.
\end{itemize}

Simultaneous dual-arm motions use an asymmetric \textit{leader-follower} setup: Arm~A follows the planned, while Arm~B tracks a reference via a reactive \texttt{mink} QP with collision-avoidance constraints. We have used mink here due to its natural incoporation with mujoco and being a QP solver it can incorporate posture tasks (for warm starts), collision handling and velocity/joint limits in its optimization while finding optimal control steps to control the arm (during planning we wanted speed in order to iterate fast upon a lot of possible configuration and solutions, thus used EAIK). Arm~A's joints are pinned, and transport segments are executed as constrained joint tracking to faithfully follow the object trajectory.

\subsection{Task Design}
\label{sec:task_design}
The tasks are designed with two objectives: (1) to enforce heterogeneous gripper use by making at least one trajectory segment infeasible for a single arm, requiring a handover; and (2) to introduce reachability-based infeasibility, demonstrating that the framework generalizes beyond collision constraints. We design both single-switch and two-switch tasks, described below (see Figure~\ref{fig:tasks} for task examples).

\begin{itemize}
    \item \textbf{Task A: Plate from Table to Shelf} A flat plate is positioned flush on the table to mandate a suction pick, while the destination shelf's narrow clearance prohibits suction entry, deliberately forcing a handover for final placement.
    \item \textbf{Task B: YCB Plate from Shelf to Table} The recessed YCB plate blocks suction access, requiring a parallel-jaw grasp; however, tight boundaries at the target cause gripper collisions, necessitating a handover to suction for top-down placement on the table.
    \item \textbf{Task C: Mug Inversion} An inverted mug prevents parallel-jaw rim grasping, requiring suction pickup. A handover to the parallel jaw then enables flipping and upright placement in a container.
    \item \textbf{Task D: Plate from box to box} A flat plate is transferred between two containers. Only the suction gripper can pick and place at the containers due to collision constraints for the parallel jaw. However, intermediate regions are unreachable by suction due to a wall, along the constraint that the plate can't go over it, thereby requiring two handovers. 
\end{itemize}

\begin{table*}[t]
\centering
\caption{OVERALL PERFORMANCE COMPARISON ACROSS METHODS. 
\textit{Task Success} denotes the end-to-end completion rate (assignment success $\times$ valid execution success). \textit{Handover (\%)} measures the success of valid transitions, excluding drops that occurred prior to the handover. \textit{Plan Time} is the assignment computation duration over sampled grasp trajectories, and \textit{Exec Time} is the task execution duration. \textit{Avg. Switches} reflects the mean number of arm transitions.}
\label{tab:quantitative}
\begin{tabular*}{\textwidth}{@{\extracolsep{\fill}}lcccccc}
\hline
\textbf{Method} & 
\textbf{Assign (\%)} &
\textbf{Task Success (\%)} & 
\textbf{Handover (\%)} & 
\textbf{Avg. Switches} & 
\textbf{Plan Time (s)} &
\textbf{Exec Time (s)} \\
\hline

\multicolumn{7}{c}{\textit{Single-Arm Baseline}} \\
\hline
Single Robot 
& 0.0 
& 0.0 
& - 
& 0 
& - 
& - \\

\hline
\multicolumn{7}{c}{\textit{Bimanual Method}} \\
\hline

Greedy ($beam=1$) 
& 30.0 
& 20.0 
& 75.0 
& 1.5 
& \textbf{24.03} 
& 95.8 \\

No-Diversity 
& 75.0 
& 67.5 
& \textbf{100.0} 
& \textbf{1.3} 
& 1015.87 
& \textbf{84.8} \\

No-Manipulability ($\beta=0$) 
& 80.0 
& 52.5 
& 89.1 
& 2.7 
& 502.43 
& 138.7 \\

No-Switching Penalty ($C_{\text{switch}}=0$) 
& 80.0 
& 25.0 
& 81.6 
& 10.7 
& 154.60 
& 286.5 \\

\textbf{Proposed (Exact DP)} 
& 80.0 
& \textbf{72.5} 
& \textbf{100.0} 
& \textbf{1.3} 
& 532.96 
& 92.0 \\

\textbf{Proposed (Beam 10, Stride 3)} 
& 72.5 
& 65.0 
& \textbf{100.0} 
& \textbf{1.3} 
& 91.05 
& 87.9 \\

\hline
\end{tabular*}

\vspace{0.6cm} 

\caption{PER-EXPERIMENT EVALUATION (PROPOSED: BEAM 10, STRIDE 3). 
A task-specific breakdown of the performance metrics defined in Table~\ref{tab:quantitative}.}
\label{tab:per_experiment}
\begin{tabular*}{\textwidth}{@{\extracolsep{\fill}}lcccccc}
\hline
\textbf{Task} & \textbf{Assign (\%)} & \textbf{Task Success (\%)} & \textbf{Handover (\%)} & \textbf{Avg. Switches} & \textbf{Plan Time (s)} & \textbf{Exec Time (s)} \\
\hline
Task A: Plate from table to shelf 
& 70.0 & 50.0 & 100.0 & 1.0 & 129.87 & 75.9 \\
Task B: YCB Plate from shelf to table 
& 60.0 & 60.0 & 100.0 & 1.0 & 99.37 & 81.6 \\
Task C: Mug Inversion 
& 80.0 & 80.0 & 100.0 & 1.0 & 84.67 & 63.8 \\
Task D: Plate from box to box 
& 80.0 & 70.0 & 100.0 & 2.0 & 50.30 & 127.1 \\
\hline
\end{tabular*}

\end{table*}

\subsection{Baselines}
\label{sec:baselines}
Exisiting object manipulation research with heterogeneous grippers has mainly focused on grasp generation and picking. We therefore compare our method against the following configurations:

\subsubsection{Greedy Reachability}
At each segment, the robot with a feasible grasp and higher instantaneous manipulability is greedily selected (equivalent to a beam width of 1).

\subsubsection{No-Diversity Grasp Selection}
The top-$k$ grasps are selected purely based on grasp quality scores, without applying DPP-based diversity filtering.

We also evaluated single-robot execution, which achieved 0\% success across all scenarios, confirming that the tasks require heterogeneous bimanual coordination.

\subsection{Quantitative Results}
\label{sec:quant_results}
Table~\ref{tab:quantitative} summarizes performance across 520 simulated runs to address Q1, evaluating success rates, switches, and timing.  Our exact DP method achieves \textbf{72.5\%} task success compared to the greedy baseline's 20.0\%, demonstrating the critical need for global look-ahead over myopic reachability decisions in complex transport tasks. Furthermore, excluding DPP-based grasp diversity (67.5\% success) heavily penalizes planning time (1015.87~s), as concentrating grasps in similar poses redundantly expands the dynamic programming search space without improving workspace reachability.

While global optimization is vital, the \textbf{Beam 10, Stride 3} configuration offers a practical computational trade-off. It drastically cuts average planning time from 532.96~s (Exact DP) to \textbf{91.05~s}, while still maintaining a 65.0\% task success rate and a 100.0\% handover success rate.

\subsection{Per-Experiment Details and Evaluation}
\label{sec:per_experiment}
Table~\ref{tab:per_experiment} details performance across the four tasks using the practical Beam 10, Stride 3 configuration, maintaining a robust 100.0\% handover success globally. Task C: Mug Inversion achieved the highest success (80.0\%) and rapid execution (63.8~s). 

Conversely, Task A: Plate from Table to Shelf proved most challenging (50.0\% success, 129.87~s planning time) due to environmental complexity and increased grasp sampling (60 vs. 40). Failures primarily stemmed from the inability to find valid robot assignments or confined insertion tolerances that forced suboptimal grasps, leading to transport slippage. Tasks B and D maintained balanced rates of 60.0\% and 70.0\%.

\subsection{Ablation Studies}
\label{sec:ablations}
We evaluate the contribution of key components via ablation studies, maintaining the rest of the pipeline unchanged (Table~\ref{tab:quantitative}).

\subsubsection{Switching Penalty and Frequency ($C_{\text{switch}}=0$)}
Removing the switching penalty leads to excessive, unsystematic handovers, increasing the average switches from 1.3 to \textbf{10.7}. Consequently, task success plummets to \textbf{25.0\%}, handover success drops to \textbf{81.6\%}, and execution time increases by \textbf{211.4\%}, demonstrating that unregularized handovers compound failure points. Conversely, systematic switching remains highly effective: valid executed plans requiring \textbf{two switches} maintain a robust \textbf{87.0\%} task success rate, compared to \textbf{90.3\%} for \textbf{one-switch} plans. This confirms that deliberate handovers safely navigate restricted workspaces without the severe performance drops of unregularized switching.

\subsubsection{Manipulability Cost ($\beta=0$)}
Disabling the manipulability term drops overall task success to \textbf{52.5\%} and handover success to \textbf{89.1\%}. Without this objective, the solver frequently selects poorly conditioned kinematic states, resulting in jerky trajectories and unstable grasps. Prioritizing manipulability is therefore vital to smoothly and safely execute the complex, contact-rich motions required during physical dual-arm handovers.

\subsection{Real-World Validation via Sim and Real Digital Twin}
\label{sec:real_world}
To validate our simulated results, we deployed the framework on a physical dual Franka FR3 setup using Task B: YCB Plate from Shelf to Table and Task C: Mug Inversion. We bridged the sim-to-real gap via a ROS 2 and MuJoCo-based digital twin, computing dynamically verified trajectories with the Beam=10, Stride=3 configuration (which demonstrated a 60\% and 80.0\% task success rate). During physical deployment, impedance control safely absorbed minor calibration offsets during contact-rich handovers, successfully replicating the digital twin's precise and safe bimanual behavior.

\section{Limitations and Conclusion}
\label{sec:limitation_conclusion}
\textbf{Limitations} We rely on pre-computed CAD meshes for tracking and DPP-based grasp generation, which limits zero-shot deployment on novel objects. Recent advances such as FoundationPose \cite{wen2024foundationpose} have enabled on-the-fly mesh reconstruction from RGB-D observations and reduce this dependency. The framework also assumes a fixed, human-provided object trajectory. Future work could allow embodiment-aware trajectory refinement, where the planner slightly adjusts the demonstrated trajectory to achieve more stable handovers.

In conclusion, we present an embodiment-aware planning framework for heterogeneous bimanual manipulation. By jointly optimizing segment reachability, asymmetric grasp compatibility, and transition penalties, our method translates a single human demonstration into a globally optimal execution plan. Supported by extensive simulation sweeps and validated via a real-world digital twin, this approach systematically elevates simple industrial grippers toward the versatile execution of complex manipulation tasks.

\balance
\bibliographystyle{IEEEtran}
\bibliography{IEEEexample}

\end{document}